\documentclass{article}

\usepackage[final]{neurips_2021}

\usepackage[utf8]{inputenc} 
\usepackage[T1]{fontenc}    
\usepackage{hyperref}       
\usepackage{url}            
\usepackage{booktabs}       
\usepackage{amsfonts}       
\usepackage{nicefrac}       
\usepackage{microtype}      
\usepackage{xcolor}         
\usepackage{graphicx}

\usepackage{amsmath}
\usepackage{amssymb}
\usepackage{mathrsfs}
\usepackage{amsthm}

\newcommand{\cL}{\mathcal{L}}
\title{Losses, Dissonances, and Distortions}

\author{%
  Pablo Samuel Castro \\
  Google Research, Brain Team \\
  \texttt{psc@google.com} \\
}

\begin{document}

\maketitle


\section{Introduction}
In recent years, there has been a growing interest in using machine learning
models for creative purposes. In most cases, this is with the use of large
{\em generative models} which, as their name implies, can generate high-quality
and realistic outputs in music \citep{huang2018music},
images \citep{esser2021taming}, text \citep{brown2020language}, and others. The
standard approach for artistic creation using these models is to take a
{\em pre-trained} model (or set of models) and use them for producing output.
The artist directs the model's generation by ``navigating'' the latent space
\citep{castro20ganterpretations}, fine-tuning the trained parameters
\citep{dinculescu19midime}, or using the model's output to steer another
generative process \citep{white19shared,castro19performing}.

At a high-level what all these approaches are doing is converting the numerical
signal of a machine learning model's output into art, whether implicitly or
explicitly.  However, in most (if not all) cases they only do so {\em after
the initial model has been trained}. This is somewhat unfortunate, as there
are plenty of numerical signals available {\em during the training process},
such as the loss and gradient values, that can be used for creative purposes.

In this paper I present a study in using the losses and gradients obtained
during the training of a simple function approximator as a mechanism for
creating musical dissonance and visual distortion in a solo piano performance
setting. These dissonances and distortions become part of an artistic
performance not just by affecting the visualizations, but also by affecting the
artistic musical performance. The system is designed such that the performer
can in turn affect the training process itself, thereby creating a closed
feedback loop between two processes: the training of a machine learning model
and the performance of an improvised piano piece.

\section{Components}

\paragraph{Losses and gradients}
Let $f_\theta:X\rightarrow Y$ denote a function parameterized by a
$d$-dimensional vector of weights $\theta\in\mathbb{R}^d$ that aims to
approximate a ``true'' function $f:X\rightarrow Y$. We improve the
approximation by updating the parameters $\theta$ so as to minimize a loss
function $\mathcal{L}(f_\theta, f)\rightarrow\mathbb{R}$. This is typically
done using gradient descent, where we use the derivative (or gradient) of the
loss function to update the parameters:
$\theta \leftarrow \theta - \alpha \nabla \mathcal{L}(f_\theta, f)$,
where $\alpha$ is a learning rate. If set properly, this process will
result in $\cL\rightarrow 0$.

Thus, at every iteration of the learning process we have $d+1$ values
at our disposal: the $d$ partial gradients from $\nabla\cL$
and the loss itself. In the next sections I will describe how I use these
values as part of a performance, but of course there are an infinitude of
ways that artists can incorporate these as part of their work.

\begin{figure*}[!h]
  \centering
   \includegraphics[width=0.44\textwidth]{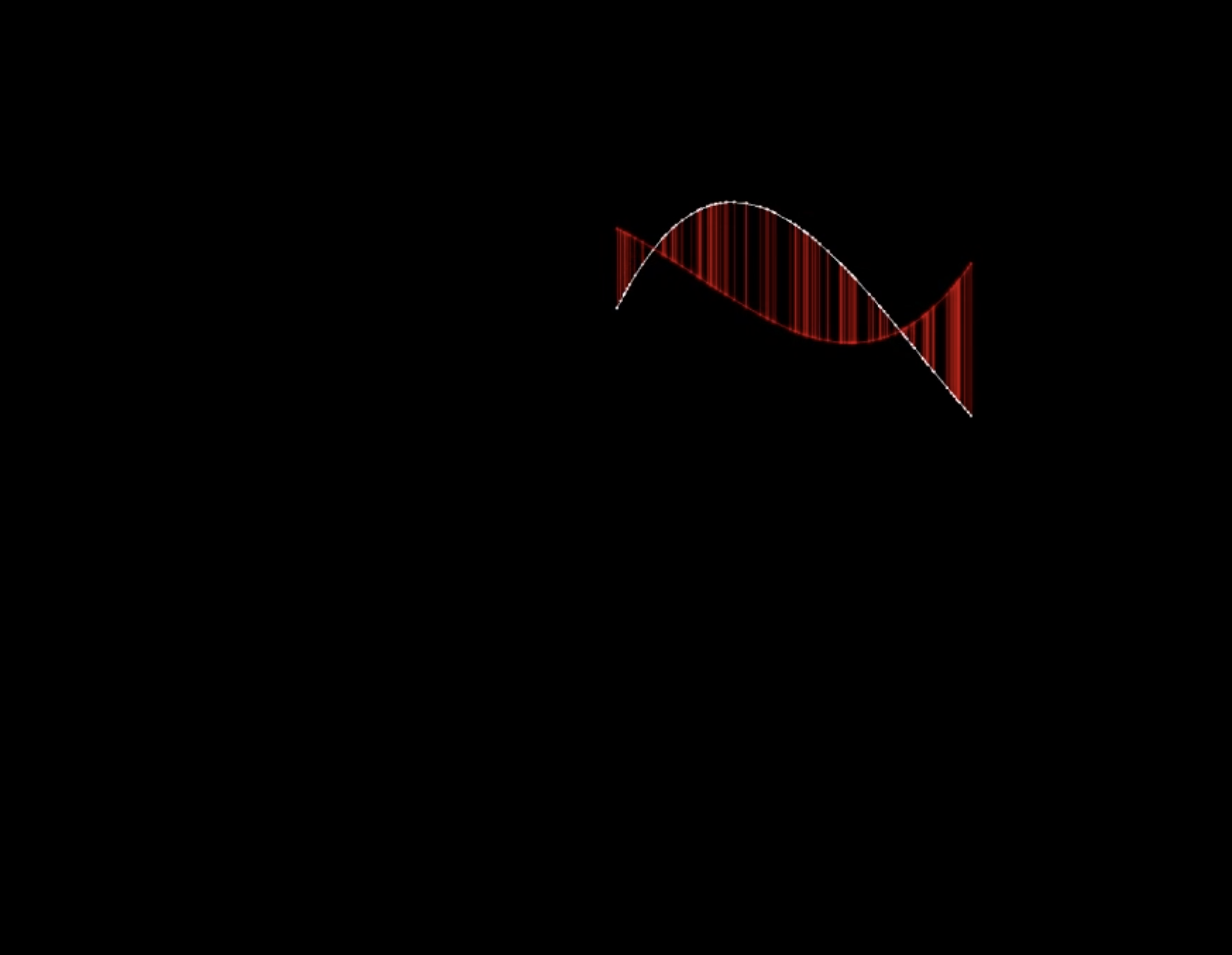}
   \includegraphics[width=0.44\textwidth]{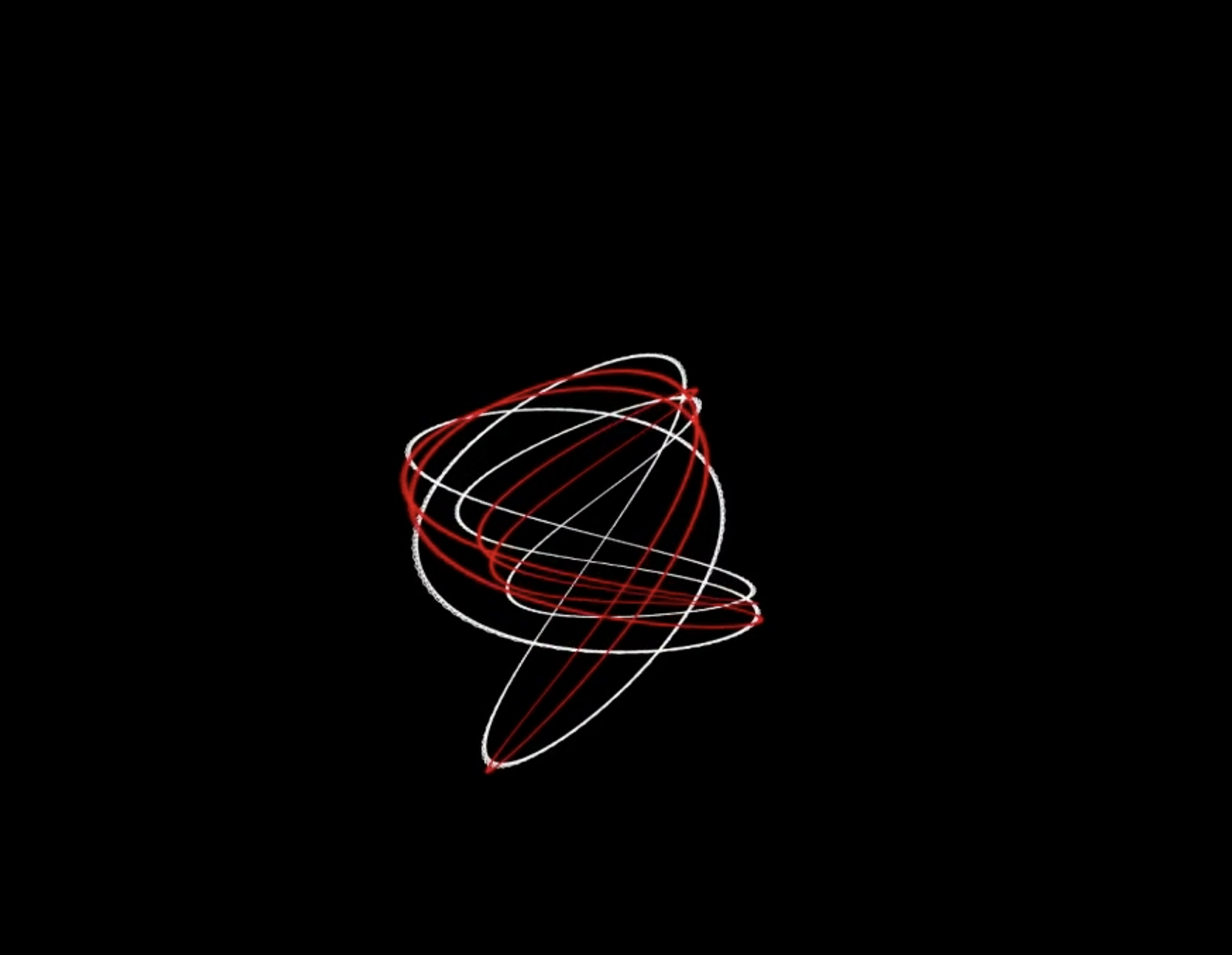}
   \includegraphics[width=0.44\textwidth]{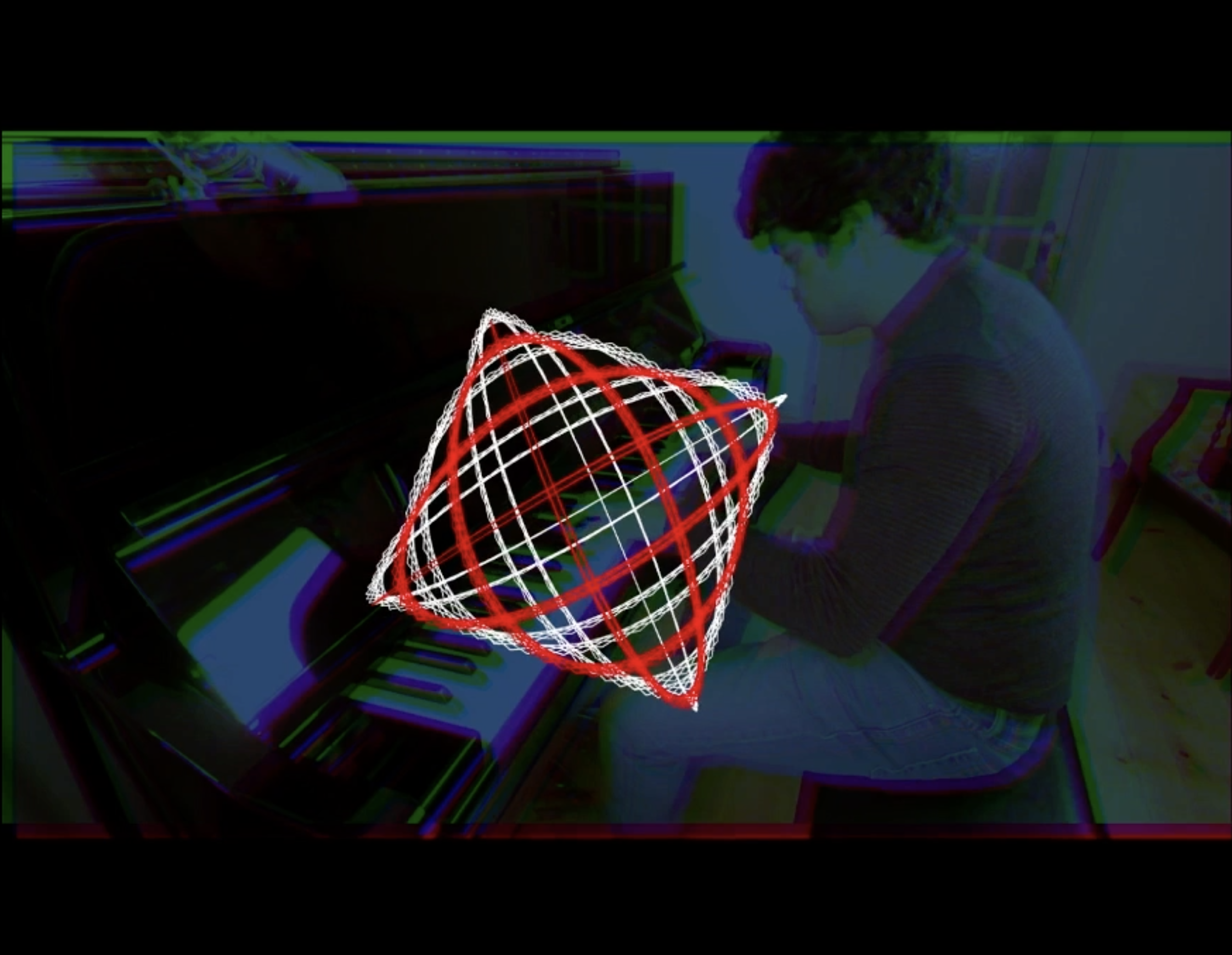}
   \includegraphics[width=0.44\textwidth]{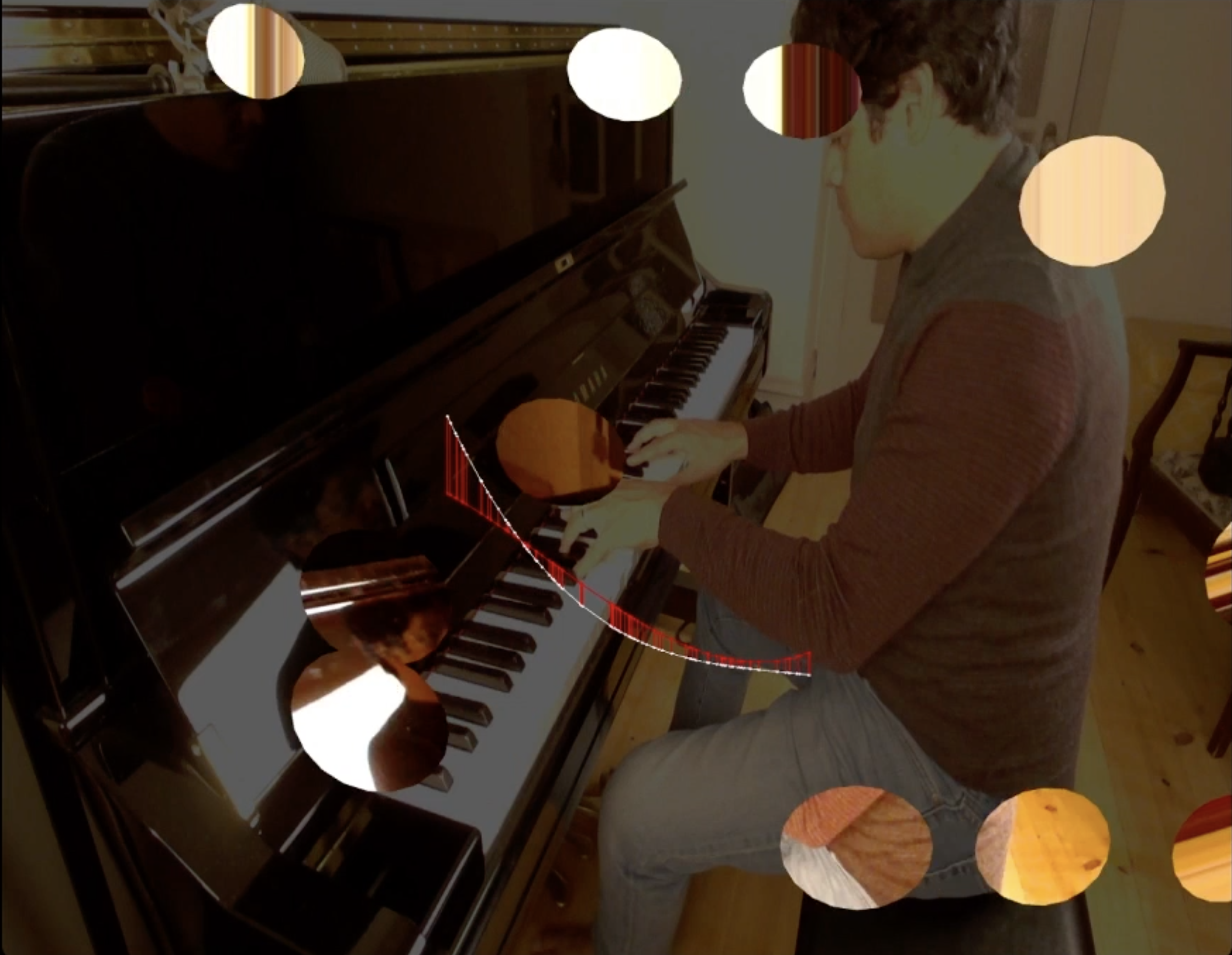}
  \caption{Screenshots of parts 1-4 (from left to right, top to bottom) of a performance. True functions are in white, learned functions are in red.}
  \label{fig:screenshots}
\end{figure*}

\paragraph{Cubics and Lissajous knots}
In order for the learning dynamics to be clearly observed during the
performance, it is preferable that the learning process is able to converge
relatively quickly. For this reason I chose two relatively simple functions to
learn: Cubic polynomials and Lissajous knots. The polynomials are single-valued
functions $f_{a,b,c,d}:\mathbb{R}\rightarrow\mathbb{R}$, while Lissajous knots
are multi-valued functions
$g_{n_x,n_y,n_z,a,b,c}:\mathbb{R}\rightarrow\mathbb{R}^3$, where $n_x$, $n_y$,
and $n_z$ are integers (see the top row of \autoref{fig:screenshots} for an
example of each):
\begin{align*}
  f_{{\bf a,b,c,d}}(x) & = ax^3 + bx^2 + cx + d \\
  g_{n_x,n_y,n_z,{\bf a,b,c}}(t) & = \langle cos(n_x t + a), cos(n_y t + b), cos(n_z t + c) \rangle
\end{align*}
In both cases, the parameters $\theta$ of the learned function aim to approach
the true values of $a$, $b$, $c$, and $d$ (e.g. the integer-parameters of the
Lissajous knots are not learned). We use the mean-squared error loss for both:
$\mathbb{E}_x\left[\sqrt{\left(f_{a,b,c,d}(x) - f_\theta\right(x))^2}\right]$ and
$\mathbb{E}_t\left[\sqrt{\left(g_{n_x,n_y,n_z,a,b,c}(t) - g_\theta\right(t))^2}\right]$.

\paragraph{Dissonances}
Music is made of the combination of individual {\em notes} played on a variety
of instruments. Each note is actually a combination of a number of pitches or
frequencies: the {\em fundamental frequency}\footnote{This is typically what
is referred to as "the pitch" of a played note.}; and a series of {\em overtone
frequencies}, that are pitches at higher frequencies than the fundamental.
A well-tuned instrument will have overtones that are {\em multiples} of the
fundamental frequency (and in this case, these are called harmonics). For
example, a well-tuned A note may have the following frequencies (one
fundamental and three overtones): $\{ 440, 880, 1320, 1760 \}$.
If we detune the overtones by an amount proportional to the loss:
$\{ 440, 880 (1 + \cL), 1320 (1 + \cL), 1760 (1 + \cL) \}$, then what
we will hear throughout the learning process is the sound ``converging''
to its well-tuned state, starting from a detuned state.

\paragraph{Distortions}
In addition to creating dissonance, we can create visual distortions using the
partial gradients of $\nabla\cL$, and two instances of this are explored: \\
{\bf 1)} The video input is split into its RGB components and each is
translated by an amount proportional to the first three partial gradients
of $\nabla\cL$. Thus, when fully converged, each of these gradients will be
zero, and each of the three RGB frames will be exactly superimposed, resulting
in an unaltered image. \\
{\bf 2)} The previous distortion distorted the placement of the
RGB components but kept the aspect ratios of each unaltered. In this distortion
the RGB components are unaltered, but the $(x,y)$ positions of
each pixel are distorted by an amount proportional to $(cos(\nabla\cL_1), cos(\nabla\cL_2))$,
where $\nabla\cL_i$ denotes the $i$-th partial derivative of $\nabla\cL$.

\section{Performance and Conclusion}
The above ideas are combined into a musical performance, played on a Disklavier
piano, which is a regular acoustic piano that can also send MIDI signal to the
computer. The performance is organized into 4 parts (see \autoref{fig:screenshots}
in the appendix for screenshots of each): \\
{\bf Part 1:} Every time a bass note is played, a new polynomial is generated
by sampling the coefficients $a,b,c,d$, and a new approximant is generated by sampling
$\theta$. Every note played on the upper half of the piano induces a gradient step, and
the loss of each step is used to detune the played note's overtones. The target and
learned polynomials are displayed on a black background.\\
{\bf Part 2:} Every time a chord is played on the left hand, a new target Lissajous
knot is generated by samplinig $n_x,n_y,n_z,a,b,c$, and a new approximant is generated
by sampling $\theta$. Gradient steps are continuously performed as long as the chord
is held, with the loss detuning the overtones of the notes being played.\\
{\bf Part 3:} Same as part 2, but we also display a video of the performer in the
background and use Distortion (1) to affect the RGB channels.\\
{\bf Part 4:} Same as part 1, but with a video of the performer in the
background. Additionally, each note played triggers a ``bubble'' superimposed on the video
which is distorted using Distortion (2).

This is meant to be an improvised ``process piece'' that is different evey time
it is performed. An example performance is provided at the following link:
https://youtu.be/Qjg0bt5hgi4.

\paragraph{Conclusion} In this work I have explored using the training
{\em process} (as opposed to just the output) of a machine learning model as
a key component of an artistic performance. Note that the models being
learned were unrelated to musical performance, yet their numerical
signals can be exploited in creative ways. As machine learning models become
more ubiquituous, I hope artists of all forms can make use of {\em all} the
available signals in artistic and creative ways.

\section{Ethical considerations}
This work uses only simulated/mathematical signals as training data, video
footage of myself, and all music performed is original. As such, there are
no real ethical concerns with this particular piece. However, as this is
also a call for artists to consider training dynamics as a key component
of their craft, we urge them to be thoughtful and respectful in their use of
machine learning models, data, and in the spirit of their expressiveness.

\bibliographystyle{plainnat}
\bibliography{ldd}

\appendix

\section{Screenshots}
Screenshots of a performance.

\section{Code details}
All the code was custom-built using SuperCollider and OpenFrameworks. Both
pieces of software communicate with each other via OSC. The code will be
made available on GitHub.

\subsection{SuperCollider}
The MIDI I/O between the computer and the Disklavier was done with code written
in SuperCollider. It is generally difficult to detune an acoustic piano
in real-time, so in addition to the regular acoustic sound of the piano, 
I superimpose synthesized sounds generated in SuperCollider,
which I can detune using the loss provided by the OpenFrameworks software.

\subsection{OpenFrameworks}
The machine learning and visualization code is written in OpenFrameworks. Because
of the simplicity of the functions being approximated, I calculate the gradients
and losses ``by hand'' instead of using a machine learning framework.

Distortion 2 is achieved by projecting the ``texture'' of the webcam onto each
bubble, and passing each of these bubbles through custom shaders (written in GLSL)
that can alter the color of each pixel and placement of each vertex.

\end{document}